%% file: main.tex
\documentclass[sigconf]{acmart}
\AtBeginDocument{%
  }
\input{packages}

\definecolor{ForestGreen}{RGB}{34,139,34}

\setcopyright{none}
\renewcommand\footnotetextcopyrightpermission[1]{} 
\acmConference[]{2nd HEAL Workshop at CHI Conference on Human Factors in Computing Systems}{April 26,
  2025}{Yokohama, Japan}

\begin{document}

\title[Meta-Evaluating Local LLMs]
{Meta-Evaluating Local LLMs: Rethinking Performance Metrics for Serious Games}

\input{Figures/Figure_game}
\input{Text/00_authors}

\input{Text/00_abstract}

\input{Text/00_CCSXML}

\input{Text/00_keywords}
  


\maketitle

\input{Text/00_structure}

\begin{acks}
This research was funded by the Portuguese Recovery and Resilience Program (PRR), IAPMEI/ANI/FCT under Agenda no.26, C645022399-00000057 (eGamesLab). The authors would also like to acknowledge the Portuguese Foundation for Science and Technology for projects 10.54499/LA/P/0083/2020; 10.54499/UIDP/50009/2020 \& 10.54499/UIDB/50009/2020. LP acknowledges FCT for the project CEECIND/01179/2017.
\end{acks}
\newpage
\bibliographystyle{ACM-Reference-Format}
\bibliography{References/main}


\end{document}

%% file: packages.tex
\usepackage[acronym]{glossaries}
\input{acronyms}
\usepackage{tabularx}
\usepackage{multirow}
\usepackage{booktabs}
\newcolumntype{Y}{>{\centering\arraybackslash}X}
\usepackage{tikz}
\usepackage{colortbl}

%% file: acronyms.tex
\newacronym[\glslongpluralkey={Energy Communities}]{EC}{EC}{Energy Communities}
\newacronym{LLM}{LLM}{Large Language Model}

\newacronym{TP}{TP}{True Positive}
\newacronym{TN}{TN}{True Negative}
\newacronym{FP}{FP}{False Positive}
\newacronym{FN}{FN}{False Negative}

\newacronym{TPR}{TPR}{True Positive Rate}
\newacronym{FPR}{TPR}{False Positive Rate}
\newacronym{TNR}{TPR}{True Negative Rate}
\newacronym{FNR}{TPR}{False Negative Rate}

\newacronym{PPV}{PPV}{Positive Predictive Value}
\newacronym{NPV}{NPV}{Negative Predictive Value}

%% file: Figures/Figure_game.tex
\begin{teaserfigure}
    \centering
    \includegraphics[width=1\textwidth]{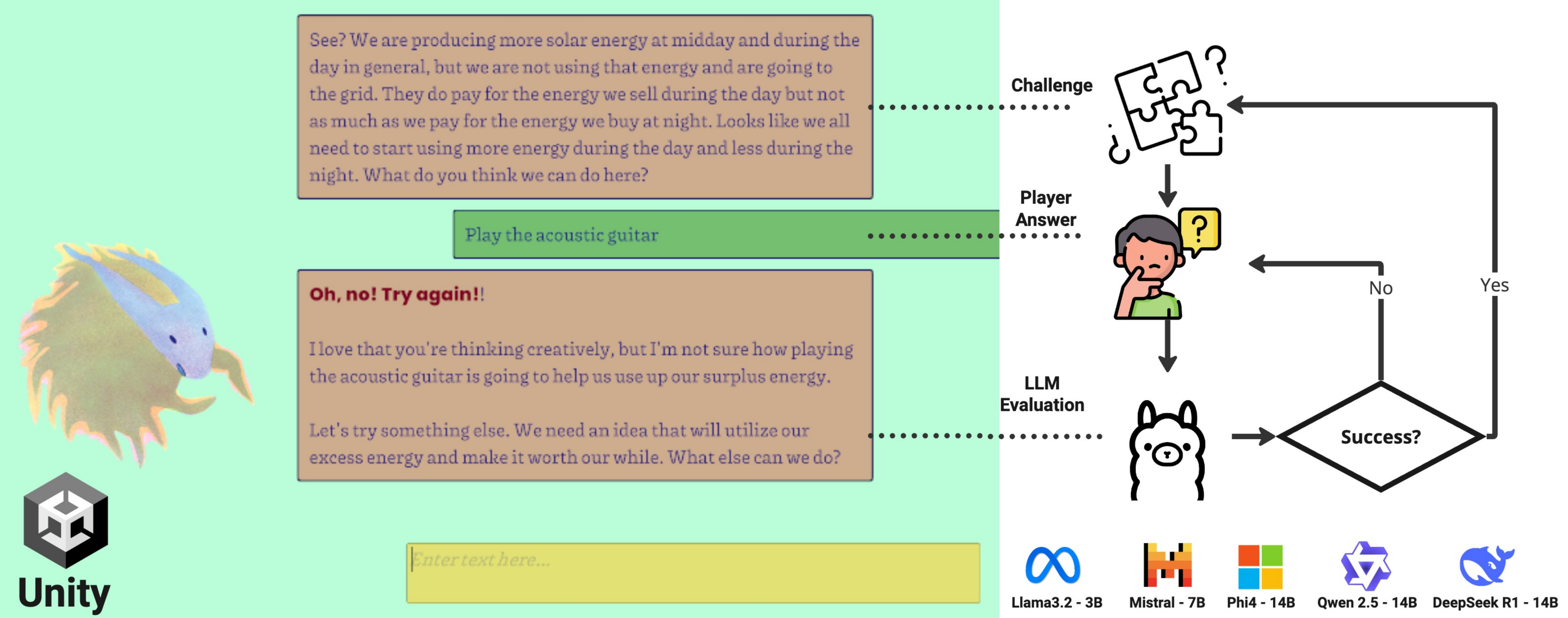}
    \caption{ In \textit{En-join} gameplay, an energy community challenge is presented to a player; the player's response is evaluated by a local LLM; as the player progresses, challenges increase in complexity.}
    \label{fig:game}
     \Description{}
\end{teaserfigure}

%% file: Text/00_authors.tex
\author{Andrés Isaza-Giraldo}
\orcid{0009-0003-9054-1491}
\email{isazag@gmail.com}
\affiliation{%
  \institution{ITI/LARSyS}
  \city{Lisbon}
  \country{Portugal}}

\author{Paulo Bala}
\email{paulo.bala@tecnico.ulisboa.pt}
\orcid{0000-0003-1743-0261}
\affiliation{%
  \institution{ITI/LARSyS, Técnico Lisboa}
  \city{Lisbon}
  \country{Portugal}}

 \author{Lucas Pereira}
\email{lucas.pereira@tecnico.ulisboa.pt}
\orcid{0000-0002-9110-8775}
\affiliation{%
  \institution{ITI/LARSyS, Técnico Lisboa}
  \city{Lisbon}
  \country{Portugal},}


%% file: Text/00_abstract.tex
\begin{abstract}

The evaluation of open-ended responses in serious games presents a unique challenge, as correctness is often subjective. Large Language Models (LLMs) are increasingly being explored as evaluators in such contexts, yet their accuracy and consistency remain uncertain, particularly for smaller models intended for local execution. This study investigates the reliability of five small-scale LLMs when assessing player responses in \textit{En-join}, a game that simulates decision-making within energy communities. By leveraging traditional binary classification metrics (including accuracy, true positive rate, and true negative rate), we systematically compare these models across different evaluation scenarios. Our results highlight the strengths and limitations of each model, revealing trade-offs between sensitivity, specificity, and overall performance. We demonstrate that while some models excel at identifying correct responses, others struggle with false positives or inconsistent evaluations. The findings highlight the need for context-aware evaluation frameworks and careful model selection when deploying LLMs as evaluators. This work contributes to the broader discourse on the trustworthiness of AI-driven assessment tools, offering insights into how different LLM architectures handle subjective evaluation tasks.

\end{abstract}

%% file: Text/00_CCSXML.tex
\ccsdesc[500]{Computing methodologies~Artificial intelligence}

%% file: Text/00_keywords.tex
\keywords{Large Language Models, Energy Communities, Environmental Games, Serious Games, LLM-as-judge}

%% file: Text/00_structure.tex
\input{Text/01_Introduction}

\input{Text/02_State_of_the_Art}

\input{Text/03_Game}

\input{Text/04_Methodology}

\input{Text/05_Results}

\input{Text/06_Conclusion}

%% file: Text/01_Introduction.tex
\section{Introduction}

The increasing capabilities of Large Language Models (LLMs) have led to their adoption across various domains, including game-based learning \cite{yang2024gptgamesupdatedscoping,sweetser_large_2024} and evaluation tasks \cite{li_generation_2025}. However, assessing their accuracy as evaluators remains a challenge, especially when dealing with open-ended responses where correctness is subjective. In the context of serious games, where players engage in decision-making and problem-solving, LLMs have the potential to serve as evaluators by providing feedback on player responses \cite{isaza-giraldo_prompt-gaming_2024}. Yet, given the non-deterministic nature of LLMs, their judgments can vary based on their architecture, training data, and internal reasoning mechanisms.

This study investigates the accuracy (effectiveness in correctly evaluating answers) and consistency (the similarity of behavior under the same settings) of small-scale LLMs when used as evaluators in a serious game setting. Our game, \textit{En-join: We Are All in Grid Together}, presents players with challenges related to energy communities, where responses must be assessed based on their effectiveness and pro-social impact. See Fig. \ref{fig:game}. By comparing multiple LLMs of small size, for the sake of being run locally, across different challenges, we aim to determine which model provides the most reliable evaluations to be trusted for autonomous evaluation in a serious game.

To achieve this, we analyze evaluation consistency across five LLMs, using traditional binary classification metrics such as accuracy, true positive rate, and true negative rate. This approach allows us to systematically compare how different models handle subjective evaluation tasks and identify potential biases in their decision-making. Our findings contribute to the ongoing discourse on the reliability of LLM-based evaluators, offering insights into their strengths and limitations in applied contexts. 
For example, different gameplay challenges require different model capabilities depending on the complexity of the challenge, suggesting a need for mixing LLM evaluators.

Ultimately, by exploring how traditional ML metrics can be applied to assess the performance and limitations of LLMs, this work seeks to trigger the discussion and contribute to establishing human-centered evaluation methods and tools for LLM-based serious games.

%% file: Text/02_State_of_the_Art.tex
\section{Related Work}

Several studies have investigated the utilization of LLMs as evaluators, also known as LLM-as-judge, suggesting a competitive correlation with human judgment \cite{wang2023chatgpt,kojima2023large,isaza-giraldo_prompt-gaming_2024,chern_can_2024,li_generation_2025,murugadoss_evaluating_2024,Wang2024Survey}. Due to the novelty of LLM systems in the field, there are not defined standardized methods for evaluating complex language behavior as shown by scoping reviews such as evaluation metrics to evaluate LLM humanoid behavior \cite{chen_humanoid} and scoping review on evaluation of LLM-as-judge capabilities \cite{li_generation_2025}. Schroeder et al. suggest that evaluation characteristics of LLMs are somehow unreliable and might change due to factors such as temperature \cite{schroeder_can_2024}. Furthermore, it was found by these researchers that there might be an inverse correlation between conversational performance and evaluation reliability \cite{schroeder_can_2024}. Because the serious game proposed implements agents that play nonplayer characters (NPC) while simultaneously performing evaluation, we found it necessary to validate the reliability of the evaluation of popular small-size LLMs as conversational-evaluating agents in the game.

Some studies have used ground-truth to evaluate model responses, Zhou et al. suggesting that LLMs are not yet reliable for reviewing long and complex papers \cite{zhou-etal-2024-llm}, and Fu et al. found that LLMs are unreliable for faculty evaluation \cite{fu_are_2023}. The use of ground-truth for evaluating LLMs zero-shot capabilities is constrained by the availability of trusted evaluation data. In another study, Lim et al. found that one LLM was very good at identifying logical fallacies based on an existing dataset \cite{lim_evaluation}, although the authors acknowledge that their results are constrained by available data. We believe that there is no guarantee that previously existing datasets are not part of the training data of any LLM, rendering this approach unsuitable for zero-shot examination. It is suggested by the authors to make evaluations of any LLM that is used for a specific task before implementation. 

%% file: Text/03_Game.tex
\section{Game Design}

\input{Figures/Figure_game}
We are developing a game, \textit{En-join}, that simulates negotiation with EC members by the introduction of open-ended challenges. The challenges have been previously crafted as authorial text and the player has to provide a written answer. The agent evaluates the player's answer determining if the provided answer is effective and pro-social. This is the only indication of evaluation that the agent receives in its prompt, and thus the evaluation depends on the zero-shot capabilities of the LLM used. The agent is instructed to reply to the player starting with either "Success!" or "Fail" and follow with a conversational answer. Normally, the model follows with a brief justification of why the evaluation was provided. In the case of negative evaluations, it was observed that the agent would make suggestions or give hints. The base prompt given to the agent is: 

\begin{quote}
\textit{You are a conversational character in an energy community game, during the level you will take the described personality to chat with the player. Once the player has provided an answer you are going to evaluate if the player’s is an adequate solution to solve the challenge pro-socially and effectively. If it is an adequate solution, start a message with 'Success!'. If it is not an adequate solution, start the message with 'Fail'. Always start the message with either 'Success!' or 'Fail'. Keep answers short.}
\end{quote}

This is followed by a "Character Description:" and the "Initial Text", which is an authorial text that composes the challenge and is presented to the player. Out of the 9 levels of the game, a sample of 3 levels was chosen for this experiment to represent ECs of different sizes, characters of different degrees of relationship, and types of challenge. Level 1 corresponds to the character "Partner", is a single household EC and the challenge is balancing energy consumption to meet solar energy production; level 2 corresponds to the character "Old Baker", is a neighborhood EC and the challenge is dealing with a neighbor that forgets its energy time slots; and level 3 corresponds to the character "Mayor Ray", is a city-wide grid and the challenge is managing AC simultaneous loads in summer. The authorial challenges for each level are described in Table \ref{tab:challenges}.

\begin{table}[!h]
    \centering
    \caption{Energy Community Challenges per level}
    \renewcommand{\arraystretch}{1} 
    \begin{tabular}{l|p{7cm}} 
        Level & Challenge \\
        \toprule
        L1 & Do you see my love? Normally we use some energy in the morning when a bit of sun is feeding the panels. Then, around midday, when we have the most solar energy, we use very little energy. Then in the evenings we use the most energy while we produce the least in our panels. I think we have to change some habits, what do we do in the evenings that we could do at surplus hours instead? \\ 
        \midrule
        L2 & Hello, my dear neighbor. You see, the energy has been so unstable lately. Yesterday I lost a whole batch of dough. Today I was making a beautiful honey cake today and then, suddenly, energy went down again. I have my batter ready to bake but my oven is not working. Have I forgotten my energy schedule again? I want to help the community but I always forget my time schedule, what could I do to avoid this happening again? \\
        \midrule
        L3 & Greeting citizen. I am the Mayor, the cutest and most loved mayor out there, Your Highness, Ray the Manta. I have a mission for you. Remember those clouds? GONE! You know what now? Too much sun? Well, not bad for your panels but you see, everybody is using their air conditioners all day long because it is so freaking hot and now we are consuming so much energy that there are outages AGAIN! IT'S HOT YOU SEE! What should I propose to the city council to solve that everybody is using their ACs at the same time? \\ 
        \bottomrule
    \end{tabular}
    \label{tab:challenges}
\end{table}

%% file: Text/04_Methodology.tex
\section{Study Design}

To understand the behavior of LLMs for the game \textit{En-join} we tested different small LLMs that can be run locally using Ollama. We selected three different challenges, as shown in Table \ref{tab:challenges}, and we tested them with the six answers shown in Fig. \ref{fig:answerseval} designed for this experiment. The output evaluation of the model was analyzed using quantitative methods based on machine learning metrics, and some qualitative analysis was performed by analyzing the model answers.

\color{black}

\subsection{Tested Answers}

\input{Figures/Figure_answers}

For each level, six answers (see fig. \ref{fig:answerseval}) were designed to assess the model's evaluation capabilities within a controlled environment while incorporating diverse answer types. The validity of the answers, i.e., the ground truth, was assessed by an expert in the energy field, including energy communities.

The answers were designed in hand with an expert on energy to render a balanced dataset, with an equal number of answers expected to be evaluated as true or false \cite{andres_CHI_2025_cr}. Specifically, half of the answers fall into the false category: one that is evidently false or ridiculous, one that is false due to failing to solve the problem, and one that lacks sufficient specific information to determine its correctness. The other half falls into the true category: one that is effective but not prosocial, one that is prosocial but not effective, and one that is both prosocial and effective. This structured distribution ensures that the evaluation does not favor either class, making the classification task evenly distributed. 

\subsection{Tested LLMs}
We evaluated several \glspl{LLM}, which were selected due to the variation in their sizes and training corpora. Llama 3.2 - 3B \cite{llama3herdmodels}, developed by Meta, is available in various parameter sizes, including 3-billion models, and is optimized for multilingual dialogue tasks, such as agentic retrieval and summarization. Mistral - 7B \cite{mistral7b}, from Mistral AI, includes models like Mistral Small 3, which is designed to be efficient, allowing operation on consumer-grade hardware while maintaining competitive performance. Phi-4 \cite{phi4technicalreport}, developed by Microsoft, is a 14-billion parameter model trained with an emphasis on high-quality data, incorporating synthetic data generation techniques to enhance reasoning and problem-solving capabilities. Qwen 2.5 \cite{qwen2technicalreport}, from Alibaba, offers models ranging from 0.5B to 72B parameters, with the 14-billion model noted for balancing performance and accessibility, making it operable on gaming PCs with 12–24GB of VRAM. Finally, DeepSeek R1 \cite{deepseekai2025deepseekr1incentivizingreasoningcapability} is an open-source AI model trained from scratch on a massive dataset of 2 trillion tokens in both English and Chinese, designed to enhance reasoning and data retrieval tasks with an option for 14-billion parameters. These models represent a diverse range of architectures and training methodologies, reflecting the rapid advancements in the field of large language models. All models were run with the default temperature of 0.8 in 14-billion parameters configuration, except for Llama 3.2, which was run on 3-billion parameters, and Mistral with 7-billion. 

\subsection{Testing Procedure}
Each answer (6 answers) was run 20 times at each level (3 levels) for each evaluator (5 LLM models), which accounted for 1800 individual evaluations. 
All the answers can be that can be consulted online\footnote{\url{https://paulobala.github.io/CHI2025_HEAL_Workshop/}}.

From these answers, confusion matrices were derived for each level and model. More precisely, a \gls{TP} was counted if both the expected answer and the model output were positive (analogously for \glspl{TN}). A \gls{FP} was counted when the expected answer was False and the model returned True (analogously for \glspl{FN}).

\subsection{Analysis}
To evaluate the performance of our model, we consider several key metrics derived from the confusion matrix. Table \ref{tab:metrics_1} lists the confusion matrix-based performance metrics that were used, where \textbf{Best} and \textbf{Worst} refer to the best and worst values that each metric can report.

The \gls{TPR}, also known as recall or sensitivity, measures how well the model identifies positive cases, calculated as $TP / (TP + FN)$. The \gls{TNR}, or specificity, reflects the ability to correctly classify negative cases, as is given by $TN / (TN + FP)$. The \gls{PPV}, or precision, indicates the proportion of true positive predictions among all positive predictions $(TP / (TP + FP))$, whereas the \gls{NPV} measures the proportion of true negative predictions among all negative predictions $(TN / (TN + FN))$. Finally, to provide a balanced evaluation of the model's performance, we calculated the F1 score for positives, which balances TPR and PPV $2 * (PPV * TPR) / (PPV + TPR)$, and an analogous F1 score for negatives (F1-N), which balances TNR and NPV.

\begin{table}[!h]
    \centering
    \caption{Summary of confusion matrix-based metrics.}
    \begin{tabular}{l|c|c|c}
        Metric & Symbol & Best & Worst  \\
        \toprule
        True Positive Rate & $TPR$ & 1 & 0 \\ 
        True Negative Rate & $TNR$ & 1 & 0 \\
        Accuracy & $A$ & 1 & 0 \\ 
       Positive Predictive Value & $PPV$ & 1 & 0 \\
       Negative Predictive Value & $NPV$ & 1 & 0 \\
       F1-score & $F1$ & 1 & 0 \\
       F1-negative & $F1-N$ & 1 & 0 \\
        \bottomrule
    \end{tabular}
    \label{tab:metrics_1}
\end{table}

The performance metrics assist in identifying "problem areas" AND "success areas", which is done by looking at visual patterns in the heatmap (Fig. \ref{fig:heatmap}) and the metric values (Table \ref{table:metrics}, and Table \ref{table:cm}). The analysis of the results is complemented by an assessment of the answers generated by the models, providing a more comprehensive understanding of their evaluation accuracy and decision patterns.

%% file: Figures/Figure_answers.tex
\begin{figure}
    \centering
    \includegraphics[width=\linewidth]{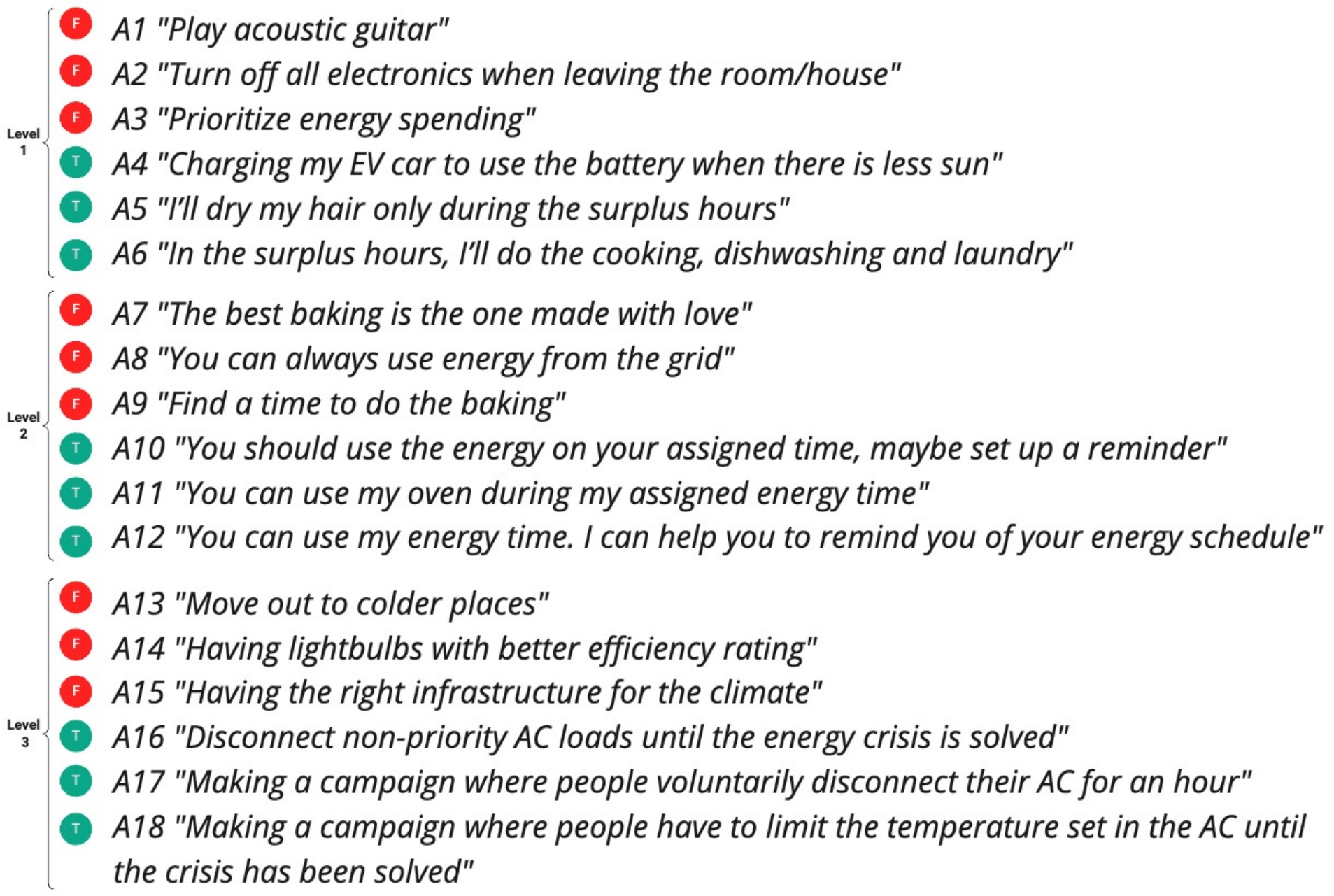}
    \caption{Tested answers. For each level, three answers are positive and three are negative.}
    \label{fig:answerseval}
    \Description{}
\end{figure}

%% file: Text/05_Results.tex
\section{Results \& Discussion}

\input{Figures/Figure_heatmap}
\input{Tables/main_table}

Our results exhibit very different behaviors from the models at different levels, as observed in Fig. \ref{fig:heatmap}. 
It was observed that Mistral, Phi-4 and Qwen exhibit fairly consistent behaviors, although only the later had an accuracy of evaluation over 70\%. Llama 3.2 and DeepSeek R1 showed a less deterministic behavior, often evaluating as T or F the same answer.

Of all models, Mistral - 7B was the most forgiving, accepting almost any answer except for A10 and A13. While it scored 90\% at evaluating positives, more than any other model, it also showed the worst results at evaluating negatives at 16\%. For a total accuracy of 53\%, it is inferior to any other model. Llama 3.2 - 3B had slightly better results at 56\% total accuracy, having the worst TPR for L3 among all tested models. Different from Mistral, its behavior was less predictable. It should be noted that these two models are the smallest of all tested, with 7 billion and 3 billion parameters, respectively. Below is an example of FP when evaluating the answer A1 ("\textit{Play acoustic guitar}"):

\begin{quote}
    \textit{Success! Playing an electric guitar requires more energy than playing an acoustic one. So, playing your acoustic guitar during surplus solar hours can help balance your energy consumption.}
\end{quote}

DeepSeek R1 had a total accuracy of 63\%, more than 10 points under models of a similar size 14 billion parameters. This model has an implemented \textit{chain of thought} that recreates a certain type of thinking behavior that analyses the factors involved in the prompt and the interaction before producing an answer. That thinking process is attached to the website along with the outputs. We believe this chain-of-thought process is the reason for a less deterministic behavior. Although this feature is very transparent and useful in other scenarios, in this case, it created a case of evaluation  where the LLM would deceive itself or end up ignoring part of the answer, as seen in A12 ("\textit{You can use my energy time. I can help you to remind you of your energy schedule}") and DeepSeek provides a fail reply after this thinking process: 

\begin{quote}
\textit{Okay, so the player responded by offering to help the Old Baker remember his energy schedule. That's a good start because it shows he's trying to be helpful and solve the problem. But wait, the challenge was about the oven not working when the energy went down again. The player suggested using his own energy time instead of fixing the root issue.}
   
\end{quote}

Qween 2.5 - 14B exhibited a relatively high overall accuracy at around 76\%. The evaluation results were fairly consistent, meaning it would provide the same evaluation in most cases. It had the best overall performance on L3 at 80\%, managing to maintain TP and TN consistently. However, it showed a high number of FN on L2, which was not seen in the other models. Below is an example of a FN when evaluating A11 ("\textit{You can use my oven during my assigned energy time}"):


\begin{quote}
    \textit{Fail The player's response does not address the immediate need of the Old Baker who requires a solution for baking his honey cake right now. Instead, it suggests using the player’s oven at a later time, which doesn't solve the current issue faced by the Old Baker. An adequate response would be to suggest alternative actions such as using an appliance that works with battery power or finding another immediate way to bake the cake within the given constraints.}
\end{quote}

Phi-4 had the best overall accuracy of all models at 79\%, having an almost perfect score on L2. It also performed well on L1, except on A1 "\textit{Play the acoustic guitar}". It seems the model was confused by the answer as it saw playing the acoustic guitar as an altruistic way to consume less energy, despite not resolving the proposed challenge that expected the player to change habits to use less energy in the evenings and more during the surplus hours. In L3, the model showed extremely negative evaluations of most answers, except for A18, for a TPR on L3 of 20\%, the second worst after Llama 3.2. However, in the evaluations, the model showed to have precise information on the topic, which might be due to the good quality of its training data. This is an example of L3, where it evaluates the answer A17 ("\textit{Making a campaign where people voluntarily disconnect their AC for an hour}"):

\begin{quote}
  \textit{Fail. While encouraging voluntary disconnection has some merit, it's unlikely to be consistent or effective enough to significantly reduce peak demand. A more robust solution would involve implementing incentives for staggered usage of air conditioning and promoting energy efficiency upgrades across the city. How about proposing a system where people can earn rebates for using their ACs during off-peak hours? Additionally, encouraging the installation of smart thermostats could help manage energy use more effectively. Let's work together to make our community cooler in every sense!}
\end{quote}

In summary, from the analysis of the model performances, it is evident that different models excel in varying aspects of classification. Phi-4 consistently demonstrates the most balanced performance across all levels, making it the ideal choice for scenarios where TPR and TNR are equally important, which is the case of the \textit{En-join} game. Qwen also performs very well, particularly regarding PPV and TNR, making it a strong candidate for use cases where minimizing false positives is a priority. Conversely, Mistral excels in TPR, detecting positive cases with high accuracy but at the expense of a high FPR. Meanwhile, Llama and DeepSeek-R1 show moderate performance, with room for improvement in consistency and TPR, particularly in level 3.

For further analysis, the individual confusion matrices for each model are presented in Table \ref{table:cm}.
\input{Tables/answers_table}

%% file: Figures/Figure_heatmap.tex
\begin{figure*}[!h]
    \centering
    \includegraphics[width=0.8\textwidth]{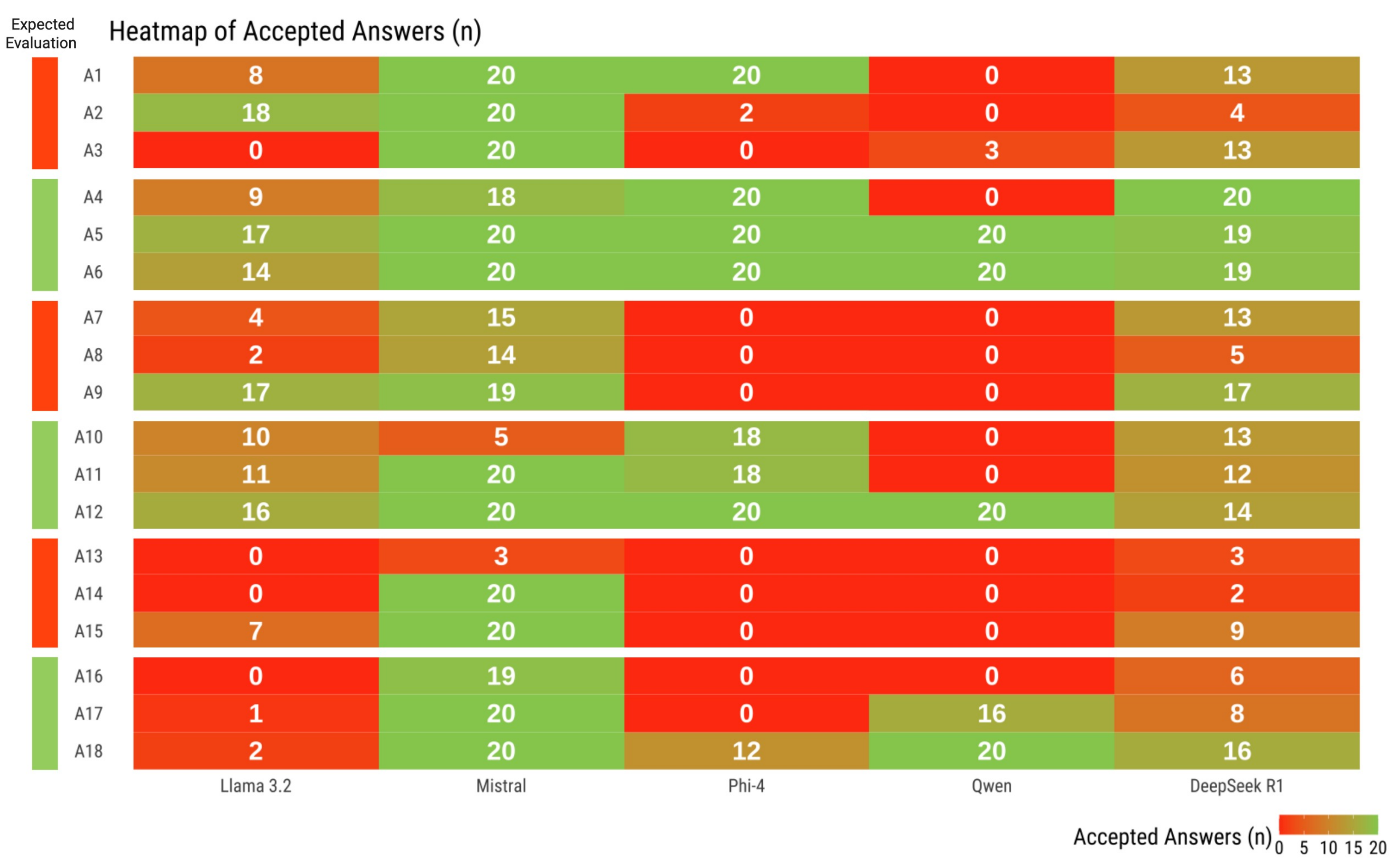}
    \caption{Heatmap of accepted answers.}
    \label{fig:heatmap}
     \Description{}
\end{figure*}

%% file: Tables/main_table.tex
\begin{table*}[!h]
\centering
\caption{Performance metric values for each LLM by level and model.}
\begin{tabular}{lc|cccc|ccccccc}
\multicolumn{1}{c}{\textbf{}} & \textbf{Level} & \textbf{TN} & \textbf{FP} & \textbf{FN} & \textbf{TP} & \textbf{TPR}  & \textbf{TNR}  & \textbf{Acc} & \multicolumn{1}{c}{\textbf{PPV}} & \multicolumn{1}{c}{\textbf{NPV}} & \multicolumn{1}{c}{\textbf{F1-P}} & \multicolumn{1}{c}{\textbf{F1-N}} \\

\toprule

Llama 3.2  & L1  & 34 & 26 & 20 & 40 & \cellcolor[HTML]{F1E784}66.67\% & \cellcolor[HTML]{FEDD81}56.67\% & \cellcolor[HTML]{FEE783}61.67\% & \cellcolor[HTML]{FEE582}60.61\% & \cellcolor[HTML]{FEEA83}62.96\% & \cellcolor[HTML]{FEE783}63.49\%  & \cellcolor[HTML]{FEDE81}59.65\%  \\
 & L2  & 37 & 23 & 23 & 37 & \cellcolor[HTML]{FEE783}61.67\% & \cellcolor[HTML]{FEE783}61.67\% & \cellcolor[HTML]{FEE783}61.67\% & \cellcolor[HTML]{FEE783}61.67\% & \cellcolor[HTML]{FEE783}61.67\% & \cellcolor[HTML]{FEE282}61.67\%  & \cellcolor[HTML]{FEE282}61.67\%  \\
 & L3  & 53 & 7  & 57 & 3  & \cellcolor[HTML]{F8736C}5.00\%  & \cellcolor[HTML]{95CD7E}88.33\% & \cellcolor[HTML]{FDC87D}46.67\% & \cellcolor[HTML]{FBA676}30.00\% & \cellcolor[HTML]{FDCB7E}48.18\% & \cellcolor[HTML]{F8696B}8.57\% & \cellcolor[HTML]{FEE482}62.35\%  \\
 & All L & 124  & 56 & 100  & 80 & \cellcolor[HTML]{FCC47C}44.44\% & \cellcolor[HTML]{E8E583}68.89\% & \cellcolor[HTML]{FEDD81}56.67\% & \cellcolor[HTML]{FEE182}58.82\% & \cellcolor[HTML]{FEDA80}55.36\% & \cellcolor[HTML]{FDC97D}50.63\%  & \cellcolor[HTML]{FEE282}61.39\%  \\ \midrule
Mistral  & L1  & 0  & 60 & 2  & 58 & \cellcolor[HTML]{72C37C}96.67\% & \cellcolor[HTML]{F8696B}0.00\%  & \cellcolor[HTML]{FDCC7E}48.33\% & \cellcolor[HTML]{FDCD7E}49.15\% & \cellcolor[HTML]{F8696B}0.00\%  & \cellcolor[HTML]{FEEA83}65.17\%  & - \\
 & L2  & 12 & 48 & 15 & 45 & \cellcolor[HTML]{CEDD82}75.00\% & \cellcolor[HTML]{FA9272}20.00\% & \cellcolor[HTML]{FDCA7D}47.50\% & \cellcolor[HTML]{FDCC7E}48.39\% & \cellcolor[HTML]{FCC47C}44.44\% & \cellcolor[HTML]{FEDC81}58.82\%  & \cellcolor[HTML]{FA9473}27.59\%  \\
 & L3  & 17 & 43 & 1  & 59 & \cellcolor[HTML]{6BC17C}98.33\% & \cellcolor[HTML]{FBA376}28.33\% & \cellcolor[HTML]{FFEB84}63.33\% & \cellcolor[HTML]{FEDF81}57.84\% & \cellcolor[HTML]{7BC57D}94.44\% & \cellcolor[HTML]{DAE182}72.84\%  & \cellcolor[HTML]{FCB97A}43.59\%  \\
 & All L & 29 & 151  & 18 & 162  & \cellcolor[HTML]{8ECB7E}90.00\% & \cellcolor[HTML]{F98A71}16.11\% & \cellcolor[HTML]{FDD57F}53.06\% & \cellcolor[HTML]{FDD37F}51.76\% & \cellcolor[HTML]{FEE783}61.70\% & \cellcolor[HTML]{FDEB84}65.72\%  & \cellcolor[HTML]{FA8F72}25.55\%  \\ \midrule
Phi-4  & L1  & 38 & 22 & 0  & 60 & \cellcolor[HTML]{63BE7B}100.00\% & \cellcolor[HTML]{FFEB84}63.33\% & \cellcolor[HTML]{B1D580}81.67\% & \cellcolor[HTML]{D6DF82}73.17\% & \cellcolor[HTML]{63BE7B}100.00\% & \cellcolor[HTML]{A0D07F}84.51\%  & \cellcolor[HTML]{C3DA81}77.55\%  \\
 & L2  & 60 & 0  & 4  & 56 & \cellcolor[HTML]{80C77D}93.33\% & \cellcolor[HTML]{63BE7B}100.00\% & \cellcolor[HTML]{72C37C}96.67\% & \cellcolor[HTML]{63BE7B}100.00\% & \cellcolor[HTML]{7EC67D}93.75\% & \cellcolor[HTML]{65BF7C}96.55\%  & \cellcolor[HTML]{63BE7B}96.77\%  \\
 & L3  & 60 & 0  & 48 & 12 & \cellcolor[HTML]{FA9272}20.00\% & \cellcolor[HTML]{63BE7B}100.00\% & \cellcolor[HTML]{FEE482}60.00\% & \cellcolor[HTML]{63BE7B}100.00\% & \cellcolor[HTML]{FEDB80}55.56\% & \cellcolor[HTML]{FBA175}33.33\%  & \cellcolor[HTML]{E1E383}71.43\%  \\
 & All L & 158  & 22 & 52 & 128  & \cellcolor[HTML]{DEE283}71.11\% & \cellcolor[HTML]{98CE7F}87.78\% & \cellcolor[HTML]{BBD881}79.44\% & \cellcolor[HTML]{A2D07F}85.33\% & \cellcolor[HTML]{CDDD82}75.24\% & \cellcolor[HTML]{C2DA81}77.58\%  & \cellcolor[HTML]{B1D580}81.03\%  \\ \midrule
Qwen 2.5 & L1  & 57 & 3  & 20 & 40 & \cellcolor[HTML]{F1E784}66.67\% & \cellcolor[HTML]{79C57D}95.00\% & \cellcolor[HTML]{B5D680}80.83\% & \cellcolor[HTML]{81C77D}93.02\% & \cellcolor[HTML]{D2DE82}74.03\% & \cellcolor[HTML]{C2DA81}77.67\%  & \cellcolor[HTML]{A7D27F}83.21\%  \\
 & L2  & 60 & 0  & 40 & 20 & \cellcolor[HTML]{FBAD78}33.33\% & \cellcolor[HTML]{63BE7B}100.00\% & \cellcolor[HTML]{F1E784}66.67\% & \cellcolor[HTML]{63BE7B}100.00\% & \cellcolor[HTML]{FEE482}60.00\% & \cellcolor[HTML]{FDC87D}50.00\%  & \cellcolor[HTML]{CFDE82}75.00\%  \\
 & L3  & 60 & 0  & 24 & 36 & \cellcolor[HTML]{FEE482}60.00\% & \cellcolor[HTML]{63BE7B}100.00\% & \cellcolor[HTML]{B9D780}80.00\% & \cellcolor[HTML]{63BE7B}100.00\% & \cellcolor[HTML]{DDE283}71.43\% & \cellcolor[HTML]{CFDE82}75.00\%  & \cellcolor[HTML]{A6D27F}83.33\%  \\
 & All L & 177  & 3  & 84 & 96 & \cellcolor[HTML]{FDD680}53.33\% & \cellcolor[HTML]{6BC17C}98.33\% & \cellcolor[HTML]{CADC81}75.83\% & \cellcolor[HTML]{70C27C}96.97\% & \cellcolor[HTML]{ECE683}67.82\% & \cellcolor[HTML]{EEE683}68.82\%  & \cellcolor[HTML]{B5D680}80.27\%  \\ \midrule
DeepSeek R1  & L1  & 30 & 30 & 2  & 58 & \cellcolor[HTML]{72C37C}96.67\% & \cellcolor[HTML]{FDCF7E}50.00\% & \cellcolor[HTML]{D5DF82}73.33\% & \cellcolor[HTML]{F5E884}65.91\% & \cellcolor[HTML]{7EC67D}93.75\% & \cellcolor[HTML]{BED981}78.38\%  & \cellcolor[HTML]{FFEB84}65.22\%  \\
 & L2  & 25 & 35 & 21 & 39 & \cellcolor[HTML]{F8E984}65.00\% & \cellcolor[HTML]{FCBE7B}41.67\% & \cellcolor[HTML]{FDD680}53.33\% & \cellcolor[HTML]{FDD57F}52.70\% & \cellcolor[HTML]{FED880}54.35\% & \cellcolor[HTML]{FEDA80}58.21\%  & \cellcolor[HTML]{FCC17C}47.17\%  \\
 & L3  & 46 & 14 & 30 & 30 & \cellcolor[HTML]{FDCF7E}50.00\% & \cellcolor[HTML]{C7DB81}76.67\% & \cellcolor[HTML]{FFEB84}63.33\% & \cellcolor[HTML]{EBE683}68.18\% & \cellcolor[HTML]{FEE582}60.53\% & \cellcolor[HTML]{FED980}57.69\%  & \cellcolor[HTML]{F3E884}67.65\%  \\
 & All L & 101  & 79 & 53 & 127  & \cellcolor[HTML]{E1E383}70.56\% & \cellcolor[HTML]{FEDC81}56.11\% & \cellcolor[HTML]{FFEB84}63.33\% & \cellcolor[HTML]{FEE783}61.65\% & \cellcolor[HTML]{F6E984}65.58\% & \cellcolor[HTML]{FDEB84}65.80\%  & \cellcolor[HTML]{FEE081}60.48\% 

\end{tabular}
\label{table:metrics}
\end{table*}

%% file: Tables/answers_table.tex
\begin{table*}[!h]
\centering
\caption{Confusion Matrix for each LLM by Answer.}
\begin{tabular}{l|llll|llll|llll|llll|llll}
\multicolumn{1}{c|}{\textbf{}}       & \multicolumn{4}{c|}{\textbf{Llama 3.2}}       & \multicolumn{4}{c|}{\textbf{Mistral}}     & \multicolumn{4}{c|}{\textbf{Phi-4}}       & \multicolumn{4}{c|}{\textbf{Qwen 2.5}}        & \multicolumn{4}{c}{\textbf{DeepSeek R1}}    \\
\multicolumn{1}{c|}{\textbf{Answer}} & 
\multicolumn{1}{c}{\textbf{TN}} & \multicolumn{1}{c}{\textbf{FP}} & \multicolumn{1}{c}{\textbf{FN}} & \multicolumn{1}{c|}{\textbf{TP}} & 
\multicolumn{1}{c}{\textbf{TN}} & \multicolumn{1}{c}{\textbf{FP}} & \multicolumn{1}{c}{\textbf{FN}} & \multicolumn{1}{c|}{\textbf{TP}} & 
\multicolumn{1}{c}{\textbf{TN}} & \multicolumn{1}{c}{\textbf{FP}} & \multicolumn{1}{c}{\textbf{FN}} & \multicolumn{1}{c|}{\textbf{TP}} & \multicolumn{1}{c}{\textbf{TN}} & \multicolumn{1}{c}{\textbf{FP}} & \multicolumn{1}{c}{\textbf{FN}} & \multicolumn{1}{c|}{\textbf{TP}} & 
\multicolumn{1}{c}{\textbf{TN}} & \multicolumn{1}{c}{\textbf{FP}} & \multicolumn{1}{c}{\textbf{FN}} & \multicolumn{1}{c}{\textbf{TP}} \\ \toprule
\tikz\draw[red,fill=red] (0,0) circle (.5ex); A1   & 12& 8& 0& 0& 0& 20& 0& 0& 0& 20& 0& 0& 20& 0& 0& 0& 7& 13& 0& 0\\
\tikz\draw[red,fill=red] (0,0) circle (.5ex); A2   & 2& 18& 0& 0& 0& 20& 0& 0& 18& 2& 0& 0& 20& 0& 0& 0& 16& 4& 0& 0\\
\tikz\draw[red,fill=red] (0,0) circle (.5ex); A3   & 20& 0& 0& 0& 0& 20& 0& 0& 20& 0& 0& 0& 17& 3& 0& 0& 7& 13& 0& 0\\
\tikz\draw[ForestGreen,fill=ForestGreen] (0,0) circle (.5ex); A4   & 0& 0& 11& 9& 0& 0& 2& 18& 0& 0& 0& 20& 0& 0& 20& 0& 0& 0& 0& 20\\
\tikz\draw[ForestGreen,fill=ForestGreen] (0,0) circle (.5ex); A5   & 0& 0& 3& 17& 0& 0& 0& 20& 0& 0& 0& 20& 0& 0& 0& 20& 0& 0& 1& 19\\
\tikz\draw[ForestGreen,fill=ForestGreen] (0,0) circle (.5ex); A6   & 0& 0& 6& 14& 0& 0& 0& 20& 0& 0& 0& 20& 0& 0& 0& 20& 0& 0& 1& 19\\
\tikz\draw[red,fill=red] (0,0) circle (.5ex); A7   & 16& 4& 0& 0& 5& 15& 0& 0& 20& 0& 0& 0& 20& 0& 0& 0& 7& 13& 0& 0\\
\tikz\draw[red,fill=red] (0,0) circle (.5ex); A8   & 18& 2& 0& 0& 6& 14& 0& 0& 20& 0& 0& 0& 20& 0& 0& 0& 15& 5& 0& 0\\
\tikz\draw[red,fill=red] (0,0) circle (.5ex); A9   & 3& 17& 0& 0& 1& 19& 0& 0& 20& 0& 0& 0& 20& 0& 0& 0& 3& 17& 0& 0\\
\tikz\draw[ForestGreen,fill=ForestGreen] (0,0) circle (.5ex); A10  & 0& 0& 10& 10& 0& 0& 15& 5& 0& 0& 2& 18& 0& 0& 20& 0& 0& 0& 7& 13\\
\tikz\draw[ForestGreen,fill=ForestGreen] (0,0) circle (.5ex); A11  & 0& 0& 9& 11& 0& 0& 0& 20& 0& 0& 2& 18& 0& 0& 20& 0& 0& 0& 8& 12\\
\tikz\draw[ForestGreen,fill=ForestGreen] (0,0) circle (.5ex); A12  & 0& 0& 4& 16& 0& 0& 0& 20& 0& 0& 0& 20& 0& 0& 0& 20& 0& 0& 6& 14\\
\tikz\draw[red,fill=red] (0,0) circle (.5ex); A13  & 20& 0& 0& 0& 17& 3& 0& 0& 20& 0& 0& 0& 20& 0& 0& 0& 17& 3& 0& 0\\
\tikz\draw[red,fill=red] (0,0) circle (.5ex); A14  & 20& 0& 0& 0& 0& 20& 0& 0& 20& 0& 0& 0& 20& 0& 0& 0& 18& 2& 0& 0\\
\tikz\draw[red,fill=red] (0,0) circle (.5ex); A15  & 13& 7& 0& 0& 0& 20& 0& 0& 20& 0& 0& 0& 20& 0& 0& 0& 11& 9& 0& 0\\
\tikz\draw[ForestGreen,fill=ForestGreen] (0,0) circle (.5ex); A16  & 0& 0& 20& 0& 0& 0& 1& 19& 0& 0& 20& 0& 0& 0& 20& 0& 0& 0& 14& 6\\
\tikz\draw[ForestGreen,fill=ForestGreen] (0,0) circle (.5ex); A17  & 0& 0& 19& 1& 0& 0& 0& 20& 0& 0& 20& 0& 0& 0& 4& 16& 0& 0& 12& 8\\
\tikz\draw[ForestGreen,fill=ForestGreen] (0,0) circle (.5ex); A18  & 0& 0& 18& 2& 0& 0& 0& 20& 0& 0& 8& 12& 0& 0& 0& 20& 0& 0& 4& 16\\ \bottomrule    
\end{tabular}
\label{table:cm}
\end{table*}

%% file: Text/06_Conclusion.tex
\section{Conclusion and Future Work}


As LLMs gain traction in the research community, particularly in HCI, it is essential to develop evaluation methods that are both interpretable and aligned with human reasoning. In this paper, we explore using traditional machine learning metrics—such as TPR, TNR, Accuracy, and F1-Scores— as tools for human-centered evaluation and auditing. These metrics are well-established, easy to implement, and offer quantifiable insights into model performance. Importantly, they also hold the potential to correlate with human judgment, making them valuable for uncovering patterns and inconsistencies in LLM behavior across varying levels of complexity and task difficulty.

%


A key contribution of this study is demonstrating the inconsistency of LLMs across varying levels of task complexity within a serious games context focused on energy literacy. No single model maintained superior performance throughout, revealing important trade-offs between TPR and TNR. These inconsistencies challenge the reliability of general-purpose LLMs in domain-specific applications and emphasize the need for context-aware evaluation frameworks and adaptive tuning strategies to ensure dependable deployment in critical areas such as energy education and decision support.


In this sense, in future work, error analysis should be conducted to understand why models are more or less accurate in specific scenarios. This includes examining the outputs and decision patterns of each model to identify systematic biases or challenges in recognizing the message of a particular input. In this context, the human evaluator can play a crucial role by providing contextual insights and nuanced interpretations that automated analysis alone cannot achieve.

Additionally, one of the most promising avenues for future work involves leveraging ensemble methods to combine the strengths of Qwen and Phi-4, thus achieving a model that balances high PPV and TPR. I.e., a model that can identify all the positive cases correctly but not at the expense of an unacceptable number of FPs. This ensemble approach could use voting strategies or stacking models to optimize decision boundaries. Another potential area for exploration is the fine-tuning of models for specific levels. This involves customizing models to target performance improvement at problematic levels, such as L3 for Llama and Phi-4, where the TPR was notably low. Such fine-tuning could involve adjusting the prompt to balance PPV and TPR more effectively. In this case, the human evaluator can help adjust the prompt by providing domain-specific guidance or clarifying the intent behind ambiguous inputs. 

Finally, from a more technical perspective, it is also essential to assess the relevance of training configurations and model sizes. Evaluating how the amount and quality of training data influence model performance can reveal if certain models are underfitting (i.e., incoherent generation) or overfitting (i.e., memorization or overgeneralization). Furthermore, comparing the complexity and size of the models can help determine if performance gains are due to more parameters or more efficient architectures. This could lead to more optimized models that balance accuracy and computational efficiency.

%% file: main.bbl

\begin{thebibliography}{20}


\ifx \showCODEN    \undefined \def \showCODEN     #1{\unskip}     \fi
\ifx \showISBNx    \undefined \def \showISBNx     #1{\unskip}     \fi
\ifx \showISBNxiii \undefined \def \showISBNxiii  #1{\unskip}     \fi
\ifx \showISSN     \undefined \def \showISSN      #1{\unskip}     \fi
\ifx \showLCCN     \undefined \def \showLCCN      #1{\unskip}     \fi
\ifx \shownote     \undefined \def \shownote      #1{#1}          \fi
\ifx \showarticletitle \undefined \def \showarticletitle #1{#1}   \fi
\ifx \showURL      \undefined \def \showURL       {\relax}        \fi
\providecommand\bibfield[2]{#2}
\providecommand\bibinfo[2]{#2}
\providecommand\natexlab[1]{#1}
\providecommand\showeprint[2][]{arXiv:#2}

\bibitem[Abdin et~al\mbox{.}(2024)]%
        {phi4technicalreport}
\bibfield{author}{\bibinfo{person}{Marah Abdin}, \bibinfo{person}{Jyoti Aneja},
  \bibinfo{person}{Harkirat Behl}, \bibinfo{person}{Sébastien Bubeck},
  \bibinfo{person}{Ronen Eldan}, \bibinfo{person}{Suriya Gunasekar},
  \bibinfo{person}{Michael Harrison}, \bibinfo{person}{Russell~J. Hewett},
  \bibinfo{person}{Mojan Javaheripi}, \bibinfo{person}{Piero Kauffmann},
  \bibinfo{person}{James~R. Lee}, \bibinfo{person}{Yin~Tat Lee},
  \bibinfo{person}{Yuanzhi Li}, \bibinfo{person}{Weishung Liu},
  \bibinfo{person}{Caio C.~T. Mendes}, \bibinfo{person}{Anh Nguyen},
  \bibinfo{person}{Eric Price}, \bibinfo{person}{Gustavo de Rosa},
  \bibinfo{person}{Olli Saarikivi}, \bibinfo{person}{Adil Salim},
  \bibinfo{person}{Shital Shah}, \bibinfo{person}{Xin Wang},
  \bibinfo{person}{Rachel Ward}, \bibinfo{person}{Yue Wu},
  \bibinfo{person}{Dingli Yu}, \bibinfo{person}{Cyril Zhang}, {and}
  \bibinfo{person}{Yi Zhang}.} \bibinfo{year}{2024}\natexlab{}.
\newblock \bibinfo{title}{Phi-4 Technical Report}.
\newblock
\showeprint[arxiv]{2412.08905}~[cs.CL]
\urldef\tempurl%
\url{https://arxiv.org/abs/2412.08905}
\showURL{%
\tempurl}


\bibitem[Chen et~al\mbox{.}(2024)]%
        {chen_humanoid}
\bibfield{author}{\bibinfo{person}{Chaoran Chen}, \bibinfo{person}{Bingsheng
  Yao}, \bibinfo{person}{Yanfang Ye}, \bibinfo{person}{Dakuo Wang}, {and}
  \bibinfo{person}{Toby Jia-Jun Li}.} \bibinfo{year}{2024}\natexlab{}.
\newblock \showarticletitle{Evaluating the LLM Agents for Simulating Humanoid
  Behavior}.
\newblock \bibinfo{journal}{\emph{The First Workshop on Human-Centered
  Evaluation and Auditing of Language Models (CHI Workshop HEAL)}}
  (\bibinfo{year}{2024}).
\newblock
\urldef\tempurl%
\url{https://par.nsf.gov/biblio/10544265}
\showURL{%
\tempurl}


\bibitem[Chern et~al\mbox{.}(2024)]%
        {chern_can_2024}
\bibfield{author}{\bibinfo{person}{Steffi Chern}, \bibinfo{person}{Ethan
  Chern}, \bibinfo{person}{Graham Neubig}, {and} \bibinfo{person}{Pengfei
  Liu}.} \bibinfo{year}{2024}\natexlab{}.
\newblock \bibinfo{title}{Can Large Language Models be Trusted for Evaluation?
  Scalable Meta-Evaluation of {LLMs} as Evaluators via Agent Debate}.
\newblock
\showeprint[arxiv]{2401.16788 [cs]}
\urldef\tempurl%
\url{http://arxiv.org/abs/2401.16788}
\showURL{%
\tempurl}


\bibitem[DeepSeek-AI et~al\mbox{.}(2025)]%
        {deepseekai2025deepseekr1incentivizingreasoningcapability}
\bibfield{author}{\bibinfo{person}{DeepSeek-AI}, \bibinfo{person}{Daya Guo},
  \bibinfo{person}{Dejian Yang}, \bibinfo{person}{Haowei Zhang},
  \bibinfo{person}{Junxiao Song}, \bibinfo{person}{Ruoyu Zhang},
  \bibinfo{person}{Runxin Xu}, \bibinfo{person}{Qihao Zhu},
  \bibinfo{person}{Shirong Ma}, \bibinfo{person}{Peiyi Wang},
  \bibinfo{person}{Xiao Bi}, \bibinfo{person}{Xiaokang Zhang},
  \bibinfo{person}{Xingkai Yu}, \bibinfo{person}{Yu Wu}, \bibinfo{person}{Z.~F.
  Wu}, \bibinfo{person}{Zhibin Gou}, \bibinfo{person}{Zhihong Shao},
  \bibinfo{person}{Zhuoshu Li}, \bibinfo{person}{Ziyi Gao},
  \bibinfo{person}{Aixin Liu}, \bibinfo{person}{Bing Xue},
  \bibinfo{person}{Bingxuan Wang}, \bibinfo{person}{Bochao Wu},
  \bibinfo{person}{Bei Feng}, \bibinfo{person}{Chengda Lu},
  \bibinfo{person}{Chenggang Zhao}, \bibinfo{person}{Chengqi Deng},
  \bibinfo{person}{Chenyu Zhang}, \bibinfo{person}{Chong Ruan},
  \bibinfo{person}{Damai Dai}, \bibinfo{person}{Deli Chen},
  \bibinfo{person}{Dongjie Ji}, \bibinfo{person}{Erhang Li},
  \bibinfo{person}{Fangyun Lin}, \bibinfo{person}{Fucong Dai},
  \bibinfo{person}{Fuli Luo}, \bibinfo{person}{Guangbo Hao},
  \bibinfo{person}{Guanting Chen}, \bibinfo{person}{Guowei Li},
  \bibinfo{person}{H. Zhang}, \bibinfo{person}{Han Bao},
  \bibinfo{person}{Hanwei Xu}, \bibinfo{person}{Haocheng Wang},
  \bibinfo{person}{Honghui Ding}, \bibinfo{person}{Huajian Xin},
  \bibinfo{person}{Huazuo Gao}, \bibinfo{person}{Hui Qu}, \bibinfo{person}{Hui
  Li}, \bibinfo{person}{Jianzhong Guo}, \bibinfo{person}{Jiashi Li},
  \bibinfo{person}{Jiawei Wang}, \bibinfo{person}{Jingchang Chen},
  \bibinfo{person}{Jingyang Yuan}, \bibinfo{person}{Junjie Qiu},
  \bibinfo{person}{Junlong Li}, \bibinfo{person}{J.~L. Cai},
  \bibinfo{person}{Jiaqi Ni}, \bibinfo{person}{Jian Liang},
  \bibinfo{person}{Jin Chen}, \bibinfo{person}{Kai Dong}, \bibinfo{person}{Kai
  Hu}, \bibinfo{person}{Kaige Gao}, \bibinfo{person}{Kang Guan},
  \bibinfo{person}{Kexin Huang}, \bibinfo{person}{Kuai Yu},
  \bibinfo{person}{Lean Wang}, \bibinfo{person}{Lecong Zhang},
  \bibinfo{person}{Liang Zhao}, \bibinfo{person}{Litong Wang},
  \bibinfo{person}{Liyue Zhang}, \bibinfo{person}{Lei Xu},
  \bibinfo{person}{Leyi Xia}, \bibinfo{person}{Mingchuan Zhang},
  \bibinfo{person}{Minghua Zhang}, \bibinfo{person}{Minghui Tang},
  \bibinfo{person}{Meng Li}, \bibinfo{person}{Miaojun Wang},
  \bibinfo{person}{Mingming Li}, \bibinfo{person}{Ning Tian},
  \bibinfo{person}{Panpan Huang}, \bibinfo{person}{Peng Zhang},
  \bibinfo{person}{Qiancheng Wang}, \bibinfo{person}{Qinyu Chen},
  \bibinfo{person}{Qiushi Du}, \bibinfo{person}{Ruiqi Ge},
  \bibinfo{person}{Ruisong Zhang}, \bibinfo{person}{Ruizhe Pan},
  \bibinfo{person}{Runji Wang}, \bibinfo{person}{R.~J. Chen},
  \bibinfo{person}{R.~L. Jin}, \bibinfo{person}{Ruyi Chen},
  \bibinfo{person}{Shanghao Lu}, \bibinfo{person}{Shangyan Zhou},
  \bibinfo{person}{Shanhuang Chen}, \bibinfo{person}{Shengfeng Ye},
  \bibinfo{person}{Shiyu Wang}, \bibinfo{person}{Shuiping Yu},
  \bibinfo{person}{Shunfeng Zhou}, \bibinfo{person}{Shuting Pan},
  \bibinfo{person}{S.~S. Li}, \bibinfo{person}{Shuang Zhou},
  \bibinfo{person}{Shaoqing Wu}, \bibinfo{person}{Shengfeng Ye},
  \bibinfo{person}{Tao Yun}, \bibinfo{person}{Tian Pei},
  \bibinfo{person}{Tianyu Sun}, \bibinfo{person}{T. Wang},
  \bibinfo{person}{Wangding Zeng}, \bibinfo{person}{Wanjia Zhao},
  \bibinfo{person}{Wen Liu}, \bibinfo{person}{Wenfeng Liang},
  \bibinfo{person}{Wenjun Gao}, \bibinfo{person}{Wenqin Yu},
  \bibinfo{person}{Wentao Zhang}, \bibinfo{person}{W.~L. Xiao},
  \bibinfo{person}{Wei An}, \bibinfo{person}{Xiaodong Liu},
  \bibinfo{person}{Xiaohan Wang}, \bibinfo{person}{Xiaokang Chen},
  \bibinfo{person}{Xiaotao Nie}, \bibinfo{person}{Xin Cheng},
  \bibinfo{person}{Xin Liu}, \bibinfo{person}{Xin Xie},
  \bibinfo{person}{Xingchao Liu}, \bibinfo{person}{Xinyu Yang},
  \bibinfo{person}{Xinyuan Li}, \bibinfo{person}{Xuecheng Su},
  \bibinfo{person}{Xuheng Lin}, \bibinfo{person}{X.~Q. Li},
  \bibinfo{person}{Xiangyue Jin}, \bibinfo{person}{Xiaojin Shen},
  \bibinfo{person}{Xiaosha Chen}, \bibinfo{person}{Xiaowen Sun},
  \bibinfo{person}{Xiaoxiang Wang}, \bibinfo{person}{Xinnan Song},
  \bibinfo{person}{Xinyi Zhou}, \bibinfo{person}{Xianzu Wang},
  \bibinfo{person}{Xinxia Shan}, \bibinfo{person}{Y.~K. Li},
  \bibinfo{person}{Y.~Q. Wang}, \bibinfo{person}{Y.~X. Wei},
  \bibinfo{person}{Yang Zhang}, \bibinfo{person}{Yanhong Xu},
  \bibinfo{person}{Yao Li}, \bibinfo{person}{Yao Zhao},
  \bibinfo{person}{Yaofeng Sun}, \bibinfo{person}{Yaohui Wang},
  \bibinfo{person}{Yi Yu}, \bibinfo{person}{Yichao Zhang},
  \bibinfo{person}{Yifan Shi}, \bibinfo{person}{Yiliang Xiong},
  \bibinfo{person}{Ying He}, \bibinfo{person}{Yishi Piao},
  \bibinfo{person}{Yisong Wang}, \bibinfo{person}{Yixuan Tan},
  \bibinfo{person}{Yiyang Ma}, \bibinfo{person}{Yiyuan Liu},
  \bibinfo{person}{Yongqiang Guo}, \bibinfo{person}{Yuan Ou},
  \bibinfo{person}{Yuduan Wang}, \bibinfo{person}{Yue Gong},
  \bibinfo{person}{Yuheng Zou}, \bibinfo{person}{Yujia He},
  \bibinfo{person}{Yunfan Xiong}, \bibinfo{person}{Yuxiang Luo},
  \bibinfo{person}{Yuxiang You}, \bibinfo{person}{Yuxuan Liu},
  \bibinfo{person}{Yuyang Zhou}, \bibinfo{person}{Y.~X. Zhu},
  \bibinfo{person}{Yanhong Xu}, \bibinfo{person}{Yanping Huang},
  \bibinfo{person}{Yaohui Li}, \bibinfo{person}{Yi Zheng},
  \bibinfo{person}{Yuchen Zhu}, \bibinfo{person}{Yunxian Ma},
  \bibinfo{person}{Ying Tang}, \bibinfo{person}{Yukun Zha},
  \bibinfo{person}{Yuting Yan}, \bibinfo{person}{Z.~Z. Ren},
  \bibinfo{person}{Zehui Ren}, \bibinfo{person}{Zhangli Sha},
  \bibinfo{person}{Zhe Fu}, \bibinfo{person}{Zhean Xu}, \bibinfo{person}{Zhenda
  Xie}, \bibinfo{person}{Zhengyan Zhang}, \bibinfo{person}{Zhewen Hao},
  \bibinfo{person}{Zhicheng Ma}, \bibinfo{person}{Zhigang Yan},
  \bibinfo{person}{Zhiyu Wu}, \bibinfo{person}{Zihui Gu},
  \bibinfo{person}{Zijia Zhu}, \bibinfo{person}{Zijun Liu},
  \bibinfo{person}{Zilin Li}, \bibinfo{person}{Ziwei Xie},
  \bibinfo{person}{Ziyang Song}, \bibinfo{person}{Zizheng Pan},
  \bibinfo{person}{Zhen Huang}, \bibinfo{person}{Zhipeng Xu},
  \bibinfo{person}{Zhongyu Zhang}, {and} \bibinfo{person}{Zhen Zhang}.}
  \bibinfo{year}{2025}\natexlab{}.
\newblock \bibinfo{title}{DeepSeek-R1: Incentivizing Reasoning Capability in
  LLMs via Reinforcement Learning}.
\newblock
\showeprint[arxiv]{2501.12948}~[cs.CL]
\urldef\tempurl%
\url{https://arxiv.org/abs/2501.12948}
\showURL{%
\tempurl}


\bibitem[et~al.(2024)]%
        {llama3herdmodels}
\bibfield{author}{\bibinfo{person}{Abhimanyu~Dubey et al.}}
  \bibinfo{year}{2024}\natexlab{}.
\newblock \bibinfo{title}{The Llama 3 Herd of Models}.
\newblock
\showeprint[arxiv]{2407.21783}~[cs.AI]
\urldef\tempurl%
\url{https://arxiv.org/abs/2407.21783}
\showURL{%
\tempurl}


\bibitem[Fu et~al\mbox{.}(2023)]%
        {fu_are_2023}
\bibfield{author}{\bibinfo{person}{Xue-Yong Fu},
  \bibinfo{person}{Md~Tahmid~Rahman Laskar}, \bibinfo{person}{Cheng Chen},
  {and} \bibinfo{person}{Shashi~Bhushan {TN}}.}
  \bibinfo{year}{2023}\natexlab{}.
\newblock \bibinfo{title}{Are Large Language Models Reliable Judges? A Study on
  the Factuality Evaluation Capabilities of {LLMs}}.
\newblock
\href{https://doi.org/10.48550/arXiv.2311.00681}{doi:\nolinkurl{10.48550/arXiv.2311.00681}}
\showeprint[arxiv]{2311.00681 [cs]}


\bibitem[Isaza-Giraldo et~al\mbox{.}(2024a)]%
        {isaza-giraldo_prompt-gaming_2024}
\bibfield{author}{\bibinfo{person}{Andr\'{e}s Isaza-Giraldo},
  \bibinfo{person}{Paulo Bala}, \bibinfo{person}{Pedro~F. Campos}, {and}
  \bibinfo{person}{Lucas Pereira}.} \bibinfo{year}{2024}\natexlab{a}.
\newblock \showarticletitle{Prompt-{Gaming}: {A} {Pilot} {Study} on
  {LLM}-{Evaluating} {Agent} in a {Meaningful} {Energy} {Game}}. In
  \bibinfo{booktitle}{\emph{Extended {Abstracts} of the {CHI} {Conference} on
  {Human} {Factors} in {Computing} {Systems}}}. \bibinfo{publisher}{ACM},
  \bibinfo{address}{Honolulu HI USA}, \bibinfo{pages}{1--12}.
\newblock
\showISBNx{9798400703317}
\href{https://doi.org/10.1145/3613905.3650774}{doi:\nolinkurl{10.1145/3613905.3650774}}


\bibitem[Isaza-Giraldo et~al\mbox{.}(2024b)]%
        {andres_CHI_2025_cr}
\bibfield{author}{\bibinfo{person}{Andr\'{e}s Isaza-Giraldo},
  \bibinfo{person}{Paulo Bala}, \bibinfo{person}{Anna Jiskrov\'{a}},
  \bibinfo{person}{Luiz Sachser}, \bibinfo{person}{Pedro~F. Campos}, {and}
  \bibinfo{person}{Lucas Pereira}.} \bibinfo{year}{2024}\natexlab{b}.
\newblock \showarticletitle{Meta-evaluating the Effects of Social Preferences
  on NPC-evaluators in an Energy Community Game}. In
  \bibinfo{booktitle}{\emph{Extended {Abstracts} of the {CHI} {Conference} on
  {Human} {Factors} in {Computing} {Systems}}}. \bibinfo{publisher}{ACM},
  \bibinfo{address}{Honolulu HI USA}, \bibinfo{pages}{1--12}.
\newblock
\showISBNx{9798400703317}
\href{https://doi.org/10.1145/3613905.3650774}{doi:\nolinkurl{10.1145/3613905.3650774}}


\bibitem[Jiang et~al\mbox{.}(2023)]%
        {mistral7b}
\bibfield{author}{\bibinfo{person}{Albert~Q. Jiang}, \bibinfo{person}{Alexandre
  Sablayrolles}, \bibinfo{person}{Arthur Mensch}, \bibinfo{person}{Chris
  Bamford}, \bibinfo{person}{Devendra~Singh Chaplot}, \bibinfo{person}{Diego
  de~las Casas}, \bibinfo{person}{Florian Bressand}, \bibinfo{person}{Gianna
  Lengyel}, \bibinfo{person}{Guillaume Lample}, \bibinfo{person}{Lucile
  Saulnier}, \bibinfo{person}{Lélio~Renard Lavaud},
  \bibinfo{person}{Marie-Anne Lachaux}, \bibinfo{person}{Pierre Stock},
  \bibinfo{person}{Teven~Le Scao}, \bibinfo{person}{Thibaut Lavril},
  \bibinfo{person}{Thomas Wang}, \bibinfo{person}{Timothée Lacroix}, {and}
  \bibinfo{person}{William~El Sayed}.} \bibinfo{year}{2023}\natexlab{}.
\newblock \bibinfo{title}{Mistral 7B}.
\newblock
\showeprint[arxiv]{2310.06825}~[cs.CL]
\urldef\tempurl%
\url{https://arxiv.org/abs/2310.06825}
\showURL{%
\tempurl}


\bibitem[Kojima et~al\mbox{.}(2023)]%
        {kojima2023large}
\bibfield{author}{\bibinfo{person}{Takeshi Kojima},
  \bibinfo{person}{Shixiang~Shane Gu}, \bibinfo{person}{Machel Reid},
  \bibinfo{person}{Yutaka Matsuo}, {and} \bibinfo{person}{Yusuke Iwasawa}.}
  \bibinfo{year}{2023}\natexlab{}.
\newblock \bibinfo{title}{Large Language Models are Zero-Shot Reasoners}.
\newblock
\showeprint[arxiv]{2205.11916}~[cs.CL]


\bibitem[Li et~al\mbox{.}(2025)]%
        {li_generation_2025}
\bibfield{author}{\bibinfo{person}{Dawei Li}, \bibinfo{person}{Bohan Jiang},
  \bibinfo{person}{Liangjie Huang}, \bibinfo{person}{Alimohammad Beigi},
  \bibinfo{person}{Chengshuai Zhao}, \bibinfo{person}{Zhen Tan},
  \bibinfo{person}{Amrita Bhattacharjee}, \bibinfo{person}{Yuxuan Jiang},
  \bibinfo{person}{Canyu Chen}, \bibinfo{person}{Tianhao Wu},
  \bibinfo{person}{Kai Shu}, \bibinfo{person}{Lu Cheng}, {and}
  \bibinfo{person}{Huan Liu}.} \bibinfo{year}{2025}\natexlab{}.
\newblock \bibinfo{title}{From Generation to Judgment: Opportunities and
  Challenges of {LLM}-as-a-judge}.
\newblock
\href{https://doi.org/10.48550/arXiv.2411.16594}{doi:\nolinkurl{10.48550/arXiv.2411.16594}}
\showeprint[arxiv]{2411.16594 [cs]}


\bibitem[Lim and Perrault(2024)]%
        {lim_evaluation}
\bibfield{author}{\bibinfo{person}{Gionnieve Lim} {and}
  \bibinfo{person}{Simon~T Perrault}.} \bibinfo{year}{2024}\natexlab{}.
\newblock \showarticletitle{Evaluation of an {LLM} in Identifying Logical
  Fallacies}.
\newblock \bibinfo{journal}{\emph{The First Workshop on Human-Centered
  Evaluation and Auditing of Language Models (CHI Workshop HEAL)}}
  (\bibinfo{year}{2024}).
\newblock


\bibitem[Murugadoss et~al\mbox{.}(2025)]%
        {murugadoss_evaluating_2024}
\bibfield{author}{\bibinfo{person}{Bhuvanashree Murugadoss},
  \bibinfo{person}{Christian Poelitz}, \bibinfo{person}{Ian Drosos},
  \bibinfo{person}{Vu Le}, \bibinfo{person}{Nick {McKenna}},
  \bibinfo{person}{Carina~Suzana Negreanu}, \bibinfo{person}{Chris Parnin},
  {and} \bibinfo{person}{Advait Sarkar}.} \bibinfo{year}{2025}\natexlab{}.
\newblock \bibinfo{title}{Evaluating the Evaluator: Measuring {LLMs}' Adherence
  to Task Evaluation Instructions}.
\newblock
\href{https://doi.org/10.48550/arXiv.2408.08781}{doi:\nolinkurl{10.48550/arXiv.2408.08781}}
\showeprint[arxiv]{2408.08781 [cs]}


\bibitem[Schroeder and Wood-Doughty(2024)]%
        {schroeder_can_2024}
\bibfield{author}{\bibinfo{person}{Kayla Schroeder} {and} \bibinfo{person}{Zach
  Wood-Doughty}.} \bibinfo{year}{2024}\natexlab{}.
\newblock \bibinfo{title}{Can You Trust {LLM} Judgments? Reliability of
  {LLM}-as-a-Judge}.
\newblock
\href{https://doi.org/10.48550/arXiv.2412.12509}{doi:\nolinkurl{10.48550/arXiv.2412.12509}}
\showeprint[arxiv]{2412.12509 [cs]}


\bibitem[Sweetser(2024)]%
        {sweetser_large_2024}
\bibfield{author}{\bibinfo{person}{Penny Sweetser}.}
  \bibinfo{year}{2024}\natexlab{}.
\newblock \showarticletitle{Large {Language} {Models} and {Video} {Games}: {A}
  {Preliminary} {Scoping} {Review}}. In \bibinfo{booktitle}{\emph{{ACM}
  {Conversational} {User} {Interfaces} 2024}}. \bibinfo{publisher}{ACM},
  \bibinfo{address}{Luxembourg Luxembourg}, \bibinfo{pages}{1--8}.
\newblock
\showISBNx{9798400705113}
\href{https://doi.org/10.1145/3640794.3665582}{doi:\nolinkurl{10.1145/3640794.3665582}}


\bibitem[Wang et~al\mbox{.}(2023)]%
        {wang2023chatgpt}
\bibfield{author}{\bibinfo{person}{Jiaan Wang}, \bibinfo{person}{Yunlong
  Liang}, \bibinfo{person}{Fandong Meng}, \bibinfo{person}{Zengkui Sun},
  \bibinfo{person}{Haoxiang Shi}, \bibinfo{person}{Zhixu Li},
  \bibinfo{person}{Jinan Xu}, \bibinfo{person}{Jianfeng Qu}, {and}
  \bibinfo{person}{Jie Zhou}.} \bibinfo{year}{2023}\natexlab{}.
\newblock \bibinfo{title}{Is ChatGPT a Good NLG Evaluator? A Preliminary
  Study}.
\newblock
\showeprint[arxiv]{2303.04048}~[cs.CL]


\bibitem[Wang et~al\mbox{.}(2024)]%
        {Wang2024Survey}
\bibfield{author}{\bibinfo{person}{Lei Wang}, \bibinfo{person}{Chen Ma},
  \bibinfo{person}{Xueyang Feng}, \bibinfo{person}{Zeyu Zhang},
  \bibinfo{person}{Hao Yang}, \bibinfo{person}{Jingsen Zhang},
  \bibinfo{person}{Zhiyuan Chen}, \bibinfo{person}{Jiakai Tang},
  \bibinfo{person}{Xu Chen}, \bibinfo{person}{Yankai Lin},
  \bibinfo{person}{Wayne~Xin Zhao}, \bibinfo{person}{Zhewei Wei}, {and}
  \bibinfo{person}{Jirong Wen}.} \bibinfo{year}{2024}\natexlab{}.
\newblock \showarticletitle{A survey on large language model based autonomous
  agents}.
\newblock \bibinfo{journal}{\emph{Frontiers of Computer Science}}
  \bibinfo{volume}{18}, \bibinfo{number}{6} (\bibinfo{year}{2024}),
  \bibinfo{pages}{186345}.
\newblock
\href{https://doi.org/10.1007/s11704-024-40231-1}{doi:\nolinkurl{10.1007/s11704-024-40231-1}}


\bibitem[Yang et~al\mbox{.}(2024b)]%
        {qwen2technicalreport}
\bibfield{author}{\bibinfo{person}{An Yang}, \bibinfo{person}{Baosong Yang},
  \bibinfo{person}{Binyuan Hui}, \bibinfo{person}{Bo Zheng},
  \bibinfo{person}{Bowen Yu}, \bibinfo{person}{Chang Zhou},
  \bibinfo{person}{Chengpeng Li}, \bibinfo{person}{Chengyuan Li},
  \bibinfo{person}{Dayiheng Liu}, \bibinfo{person}{Fei Huang},
  \bibinfo{person}{Guanting Dong}, \bibinfo{person}{Haoran Wei},
  \bibinfo{person}{Huan Lin}, \bibinfo{person}{Jialong Tang},
  \bibinfo{person}{Jialin Wang}, \bibinfo{person}{Jian Yang},
  \bibinfo{person}{Jianhong Tu}, \bibinfo{person}{Jianwei Zhang},
  \bibinfo{person}{Jianxin Ma}, \bibinfo{person}{Jianxin Yang},
  \bibinfo{person}{Jin Xu}, \bibinfo{person}{Jingren Zhou},
  \bibinfo{person}{Jinze Bai}, \bibinfo{person}{Jinzheng He},
  \bibinfo{person}{Junyang Lin}, \bibinfo{person}{Kai Dang},
  \bibinfo{person}{Keming Lu}, \bibinfo{person}{Keqin Chen},
  \bibinfo{person}{Kexin Yang}, \bibinfo{person}{Mei Li},
  \bibinfo{person}{Mingfeng Xue}, \bibinfo{person}{Na Ni}, \bibinfo{person}{Pei
  Zhang}, \bibinfo{person}{Peng Wang}, \bibinfo{person}{Ru Peng},
  \bibinfo{person}{Rui Men}, \bibinfo{person}{Ruize Gao},
  \bibinfo{person}{Runji Lin}, \bibinfo{person}{Shijie Wang},
  \bibinfo{person}{Shuai Bai}, \bibinfo{person}{Sinan Tan},
  \bibinfo{person}{Tianhang Zhu}, \bibinfo{person}{Tianhao Li},
  \bibinfo{person}{Tianyu Liu}, \bibinfo{person}{Wenbin Ge},
  \bibinfo{person}{Xiaodong Deng}, \bibinfo{person}{Xiaohuan Zhou},
  \bibinfo{person}{Xingzhang Ren}, \bibinfo{person}{Xinyu Zhang},
  \bibinfo{person}{Xipin Wei}, \bibinfo{person}{Xuancheng Ren},
  \bibinfo{person}{Xuejing Liu}, \bibinfo{person}{Yang Fan},
  \bibinfo{person}{Yang Yao}, \bibinfo{person}{Yichang Zhang},
  \bibinfo{person}{Yu Wan}, \bibinfo{person}{Yunfei Chu},
  \bibinfo{person}{Yuqiong Liu}, \bibinfo{person}{Zeyu Cui},
  \bibinfo{person}{Zhenru Zhang}, \bibinfo{person}{Zhifang Guo}, {and}
  \bibinfo{person}{Zhihao Fan}.} \bibinfo{year}{2024}\natexlab{b}.
\newblock \bibinfo{title}{Qwen2 Technical Report}.
\newblock
\showeprint[arxiv]{2407.10671}~[cs.CL]
\urldef\tempurl%
\url{https://arxiv.org/abs/2407.10671}
\showURL{%
\tempurl}


\bibitem[Yang et~al\mbox{.}(2024a)]%
        {yang2024gptgamesupdatedscoping}
\bibfield{author}{\bibinfo{person}{Daijin Yang}, \bibinfo{person}{Erica
  Kleinman}, {and} \bibinfo{person}{Casper Harteveld}.}
  \bibinfo{year}{2024}\natexlab{a}.
\newblock \bibinfo{title}{GPT for Games: An Updated Scoping Review
  (2020-2024)}.
\newblock
\showeprint[arxiv]{2411.00308}~[cs.AI]
\urldef\tempurl%
\url{https://arxiv.org/abs/2411.00308}
\showURL{%
\tempurl}


\bibitem[Zhou et~al\mbox{.}(2024)]%
        {zhou-etal-2024-llm}
\bibfield{author}{\bibinfo{person}{Ruiyang Zhou}, \bibinfo{person}{Lu Chen},
  {and} \bibinfo{person}{Kai Yu}.} \bibinfo{year}{2024}\natexlab{}.
\newblock \showarticletitle{Is {LLM} a Reliable Reviewer? A Comprehensive
  Evaluation of {LLM} on Automatic Paper Reviewing Tasks}. In
  \bibinfo{booktitle}{\emph{Proceedings of the 2024 Joint International
  Conference on Computational Linguistics, Language Resources and Evaluation
  (LREC-COLING 2024)}}, \bibfield{editor}{\bibinfo{person}{Nicoletta
  Calzolari}, \bibinfo{person}{Min-Yen Kan}, \bibinfo{person}{Veronique Hoste},
  \bibinfo{person}{Alessandro Lenci}, \bibinfo{person}{Sakriani Sakti}, {and}
  \bibinfo{person}{Nianwen Xue}} (Eds.). \bibinfo{publisher}{ELRA and ICCL},
  \bibinfo{address}{Torino, Italia}, \bibinfo{pages}{9340--9351}.
\newblock
\urldef\tempurl%
\url{https://aclanthology.org/2024.lrec-main.816/}
\showURL{%
\tempurl}


\end{thebibliography}
